\documentclass[lettersize,journal]{IEEEtran}
\usepackage{amsmath,amsfonts}
\usepackage{algorithmic}
\usepackage{array}
\usepackage[caption=false,font=normalsize,labelfont=sf,textfont=sf]{subfig}
\usepackage{textcomp}
\usepackage{stfloats}
\usepackage{url}
\usepackage{verbatim}
\usepackage{graphicx}
\usepackage{cite}
\makeatletter
\newcommand{\ie}{\emph{i.e.}\@ifnextchar.{\!\@gobble}{}}
\newcommand{\eg}{\emph{e.g.}\@ifnextchar.{\!\@gobble}{}}
\newcommand{\etc}{etc\@ifnextchar.{}{.\@}}
\makeatother

\usepackage{xcolor}

\usepackage{amsmath,amsfonts,bm}

\def\eqref#1{equation~\ref{#1}}
\def\1{\bm{1}}

\def\va{{\bm{a}}}

\def\vc{{\bm{c}}}

\def\vn{{\bm{n}}}
\def\vo{{\bm{o}}}

\def\vt{{\bm{t}}}

\def\vx{{\bm{x}}}

\def\mR{{\bm{R}}}
\def\mS{{\bm{S}}}

\DeclareMathAlphabet{\mathsfit}{\encodingdefault}{\sfdefault}{m}{sl}
\SetMathAlphabet{\mathsfit}{bold}{\encodingdefault}{\sfdefault}{bx}{n}

\usepackage{amssymb}
\usepackage{multirow}
\usepackage{colortbl}  
\usepackage[ruled]{algorithm2e} % For algorithms

\SetAlFnt{\small}
\SetAlCapFnt{\small}
\SetAlCapNameFnt{\small}
\SetAlCapHSkip{0pt}
\usepackage{tablefootnote}
\usepackage{float}
\usepackage{booktabs}
\usepackage{pifont}

\usepackage[pagebackref=false,breaklinks=true,colorlinks=true,bookmarks=false,linkcolor={red!50!black},urlcolor={blue},citecolor={black}]{hyperref} 

\begin{document}

\title{GIR: 3D Gaussian Inverse Rendering for Relightable Scene Factorization}

\author{Yahao Shi, Yanmin Wu, Chenming Wu, Xing Liu, Chen Zhao, Haocheng Feng, \\ Jian Zhang, Bin Zhou, Errui Ding, Jingdong Wang
        % <-this % stops a space
\thanks{Yahao Shi and Yanmin Wu are equal contribution. (Corresponding authors: Bin Zhou and Jian Zhang).}
\thanks{Yahao Shi and Bin Zhou are with State Key Laboratory of Virtual Reality Technology and Systems, Beihang University, Beijing 100191, China (e-mail: shiyahao@buaa.edu.cn; zhoubin@buaa.edu.cn).}
\thanks{Yanmin Wu and Jian Zhang are with the School of Electronic and Computer Engineering, Peking University, Shenzhen 518055, China (e-mail: \\ wuyanminmax@gmail.com; zhangjian.sz@pku.edu.cn).}
\thanks{Chenming Wu, Xing Liu, Chen Zhao, Haocheng Feng, Errui Ding and Jingdong Wang are with VIS, Baidu Inc. (e-mail: wuchenming@baidu.com).}% <-this % stops a space
}

\markboth{Journal of \LaTeX\ Class Files,~Vol.~14, No.~8, August~2021}%
{Shell \MakeLowercase{\textit{et al.}}: A Sample Article Using IEEEtran.cls for IEEE Journals}


\maketitle
\begin{figure*}
\includegraphics[width=\textwidth]{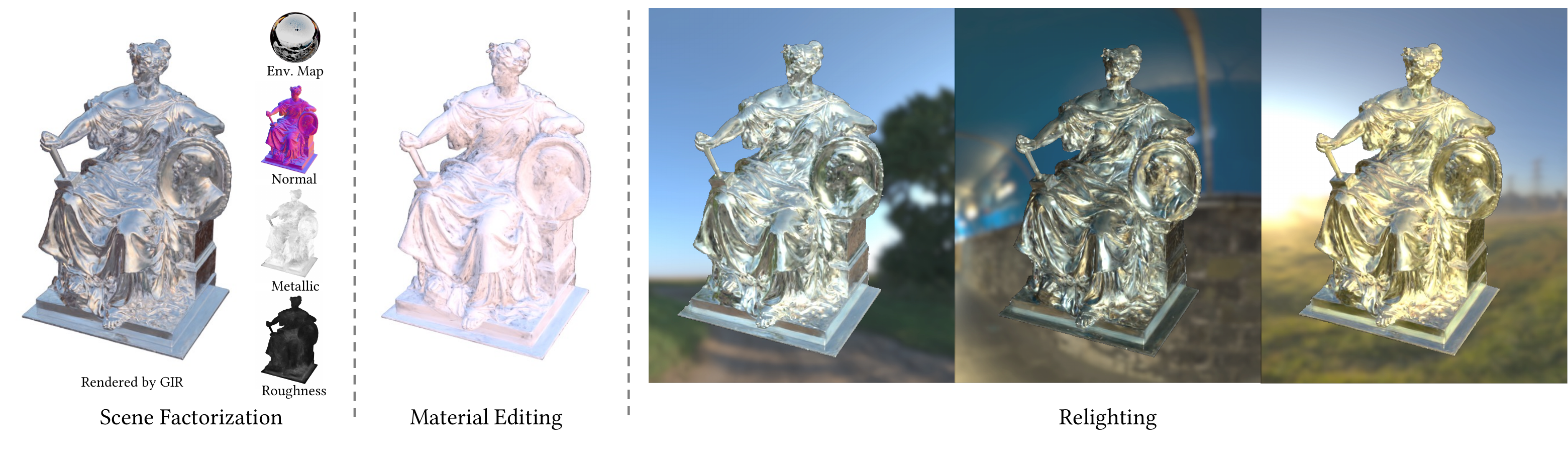}
\vspace{-26pt}
\caption{Our proposed 3D GIR offers the capability to reconstruct high-quality scenes from multi-view images, incorporating factored properties associated with physically-based rendering, thereby empowering users to interactively edit these elements or conduct relighting under novel lighting conditions.}
\vspace{-6pt}
\end{figure*}

\begin{abstract}
This paper presents a 3D Gaussian Inverse Rendering (GIR) method, employing 3D Gaussian representations to effectively factorize the scene into material properties, light, and geometry. The key contributions lie in three-fold. We compute the normal of each 3D Gaussian using the shortest eigenvector, with a directional masking scheme forcing accurate normal estimation without external supervision. We adopt an efficient voxel-based indirect illumination tracing scheme that stores direction-aware outgoing radiance in each 3D Gaussian to disentangle secondary illumination for approximating multi-bounce light transport. To further enhance the illumination disentanglement, we represent a high-resolution environmental map with a learnable low-resolution map and a lightweight, fully convolutional network. Our method achieves state-of-the-art performance in both relighting and novel view synthesis tasks among the recently proposed inverse rendering methods while achieving real-time rendering. This substantiates our proposed method's efficacy and broad applicability, highlighting its potential as an influential tool in various real-time interactive graphics applications such as material editing and relighting.
The code will be released at \href{https://github.com/guduxiaolang/GIR}{https://github.com/guduxiaolang/GIR}.
\end{abstract}

\begin{IEEEkeywords}
3D gaussian splatting, inverse rendering, relighting
%Material Editing
\end{IEEEkeywords}

\section{Introduction}
\label{sec:intro}
\IEEEPARstart{I}{nverse} rendering involves deducing scene properties, including geometry, lighting, and materials, from given images~\cite{sato1997object, marschner1998inverse, ramamoorthi2001signal, debevec2004estimating, yu1999inverse}. This longstanding challenge finds applications in various fields, such as scene understanding, image manipulation, AR/VR, etc. 
Those scene properties are critical factors in determining an object's appearance in the physical world. However, deducing scene properties exclusively from multi-view images presents an inherently ill-posed inverse challenge, attributable to the intricate integral relationships that underpin physical interactions in the real world. These relationships dictate the behavior of light transport and establish connections between geometry, materials, and illumination. Consequently, the inherent complexity of this problem inevitably introduces ambiguities in the process.

Previous efforts on neural reconstruction~\cite{munkberg2022extracting, hasselgren2022nvdiffrecmc,liu2023nero} have demonstrated effective estimation of these factors from multi-view images using implicit fields. These approaches incorporate optical rendering functions, which adhere to the principles of Bidirectional Reflectance
Distribution Function (BRDF) into trainable neural fields, enabling the estimation of physically-based rendering (PBR) components such as roughness, metallic, and environmental illuminations. In those works, discrete meshes play a significant role in conveying the factorization after the inverse rendering process is finished, because implicit fields are unfriendly for modern rendering hardware.

Recently, Kerbl et al. \cite{kerbl20233d} proposed a splatting-based approach called 3DGS. This method represents scenes using 3D Gaussians, defined by a set of properties of 3D position, opacity, anisotropic covariance, and color encoded by Spherical Harmonics (SH) coefficients. In contrast to NeRF-based approaches, which optimize implicit fields, 3DGS optimizes the explicit properties of these 3D Gaussians. 3DGS is a promising approach for tackling challenges such as dynamic scene modeling~\cite{luiten2023dynamic} and content generation~\cite{tang2023dreamgaussian}, thanks to its superior versatility and advanced performance. Furthermore, 3DGS demonstrates the remarkable potential for diverse interactive graphics applications, providing real-time rendering capabilities, competitive visual quality, and training efficiency. However, seamlessly integrating 3DGS with modern rendering engines based on PBR poses challenges since it is originally designed for novel view synthesis. Relighting under novel illuminations or material editing is intractable in the current 3DGS framework. Thus, developing an inverse rendering method upon 3DGS becomes crucial to address these limitations and unlock its full potential.

This paper introduces GIR, a 3D Gaussian Inverse Rendering method built upon 3DGS for scene factorization. Unlike previous approaches that utilize mesh or implicit field, we employ explicit 3D Gaussians to represent geometry and material properties, implementing BRDF rendering through splatting.
Incorporating inverse rendering into 3DGS poses three main challenges: 1) defining and optimizing normals within 3D Gaussians lacks a clear method. To address this, we leverage the shortest eigenvector of each 3D Gaussian to represent its normal. Additionally, we introduce a directional masking scheme to ensure accurate normals for rendering;
2) optimizing numerous parameters in 3DGS and environment maps presents difficulties, especially in solving the ill-posed inverse rendering problem. To overcome this, we design a simplified illumination representation using convolutional networks, enabling high-quality illumination disentanglement;
3) modeling indirect illumination involves ray tracing, whereas 3DGS employs splatting. We tackle this challenge by adopting an efficient voxel-based indirect illumination tracing scheme, storing direction-aware outgoing radiance within each 3D Gaussian to simulate secondary illumination. The contribution of our work can be summarized as follows.

\begin{itemize}
\item We present a novel inverse rendering framework built upon 3DGS, which enables us to estimate geometry, PBR materials, direct and indirect illuminations in a unified framework, and allows for real-time rendering.
\item We introduce a directional masking scheme for accurate normals and a simplified illumination representation for accurate environmental map estimation. Additionally, we propose an efficient voxel-based indirect illumination tracing scheme that disentangles secondary illumination.
\item Our comprehensive experiments demonstrate that our approach outperforms relevant approaches, establishing its leading performance in scene factorization.
\end{itemize}

\section{Related work}
\label{sec:related_work}
\noindent \textbf{Neural Rendering.}
Recent advances in neural rendering diverge into two strides, i.e., field-based neural rendering and primitive-based rendering.
The most representative field-based approach is NeRF~\cite{mildenhall2021nerf}, which utilizes a coordinate-based MLP to encode volumetric radiance space, enabling accurate rendering of scenes. MipNeRF~\cite{barron2021mip} proposes a multiscale representation for anti-aliasing, while Instant-NGP~\cite{muller2022instant} combines a neural network with a multiresolution hash table for efficient evaluation. ZipNeRF~\cite{barron2023zip} integrates grid-based representations and anti-aliasing for improved performance. Other approaches such as Unisurf~\cite{oechsle2021unisurf}, VolSDF~\cite{yariv2021volume}, and NeuS~\cite{wang2021neus} enhance surface reconstruction by designing proxy functions for density-to-distance field conversion, building upon NeRF's representation.

Primitive-based rendering has a well-established history that can be traced back to earlier works~\cite{csuri1979towards, levoy1985use}.
With recent neural representation advancements, Yifan et al.~\cite{yifan2019differentiable} proposes a differentiable surface splatting method for point-based geometry processing. Neural point-based graphics \cite{aliev2020neural,rakhimov2022npbg++} is a deep rendering network to generate photorealistic images by mapping rasterized point descriptors, and further improve the neural descriptors for non-Lambertian scenes without per-scene optimization. Point-NeRF~\cite{xu2022point} offers a generalizable approach using neural 3D points. ADOP~\cite{ruckert2022adop} utilizes a point cloud rasterizer with a neural network for efficient novel view generation. DeepSurfels~\cite{mihajlovic2021deepsurfels} combines geometry and appearance using explicit and neural components. Recently proposed 3DGS~\cite{kerbl20233d} introduces anisotropic 3D Gaussians and a tile-based differentiable rasterizer, advancing the field significantly. Our work embraces the principles of 3DGS while introducing the additional capability of inverse rendering for material and illumination factorization. This enhancement empowers real-time relighting and material editing, opening up new possibilities for practical applications.

\vspace{4pt} 
\noindent \textbf{Inverse Rendering.}
Inverse rendering factorizes the appearance of a scene using a collection of observed images into geometry, material, and lighting, which is a longstanding problem~\cite{sato1997object, marschner1998inverse, ramamoorthi2001signal, debevec2004estimating, yu1999inverse}. Due to its inherent under-constrained nature, learning-based approaches present more straightforward and robust to this problem.
NeRV~\cite{srinivasan2021nerv} assumes prior illumination knowledge and uses a computationally complex continuous volumetric function. NeRFactor~\cite{zhang2021nerfactor} enhances NeRF with MLPs to describe surface properties. PhySG~\cite{zhang2021physg} assumes fixed illumination conditions, while NeRD~\cite{boss2021nerd} handles both fixed and varying illumination. Neural-PIL~\cite{boss2021neural} estimates high-frequency lighting, and Ref-NeRF~\cite{verbin2022ref} models environment lighting and surface roughness. Cai et al.~\cite{cai2022physics} uses explicit-implicit hybrid representation with path tracing under natural illumination. NVDiffrec~\cite{munkberg2022extracting} uses DMTet~\cite{shen2021deep} representation with pre-integrated image-based lighting. NVDiffrecMC~\cite{hasselgren2022nvdiffrecmc} introduces path tracing and Monte Carlo integration for realistic shading but exhibits unstable training performance compared to NVDiffrec. Lyu et al.~\cite{lyu2022neural} proposes a neural radiance transfer field for precomputed global illumination with multiple-stage training with OLAT synthesis. Jin et al.~\cite{jin2023tensoir} and Mai et al.~\cite{mai2023neural} adopt a low-rank tensor structure for inverse rendering. NeILF~\cite{yao2022neilf} employs an MLP to model spatially varying illumination, capturing both direct and indirect lights. NeILF++~\cite{zhang2023neilf++} incorporates VolSDF~\cite{yariv2021volume} to learn geometry and unifies incident light and outgoing radiance through physically-based rendering. ENVIDR~\cite{liang2023envidr} addresses glossy surfaces, while NeRO~\cite{liu2023nero} focuses on reconstructing reflective objects, and similar approaches are also drawn in~\cite{mao2023neus, fan2023factored}. 

Zeng et al.~\cite{zeng2023relighting} require point light sources as input to estimate the reflectance terms decided by the networks. The number of input images is relatively high, requiring point light sources and predefined materials as priors.
Lyu et al.~\cite{lyu2023diffusion} uses path-tracing and an environment generation DDPM for disentangling illumination and material of a scene, which assumes the geometry is already obtained. Sun et al.~\cite{sun2023neural} adopts a two-stage optimization strategy, leveraging path-tracing rendering on meshes for factorization. Building upon remarkable designs from prior literature, our work showcases exceptional capabilities by enabling seamless end-to-end training and inference, eliminating the requirement for mesh conversion. This naturally leverages the advantages of image-based rendering, enabling real-time rendering, relighting, and material editing, positioning it as a noteworthy contribution to the field. Concurrent works~\cite{liang2023gs, gao2023relightable, jiang2023gaussianshader} have also successfully applied inverse rendering to 3DGS. In Sec.~\ref{sec:experiments}, we detailedly analyze the distinctions between GIR and their methods, and validate the performance of our GIR through comprehensive experiments.

\begin{figure*}[t]
\begin{center}
\includegraphics[width=\textwidth]{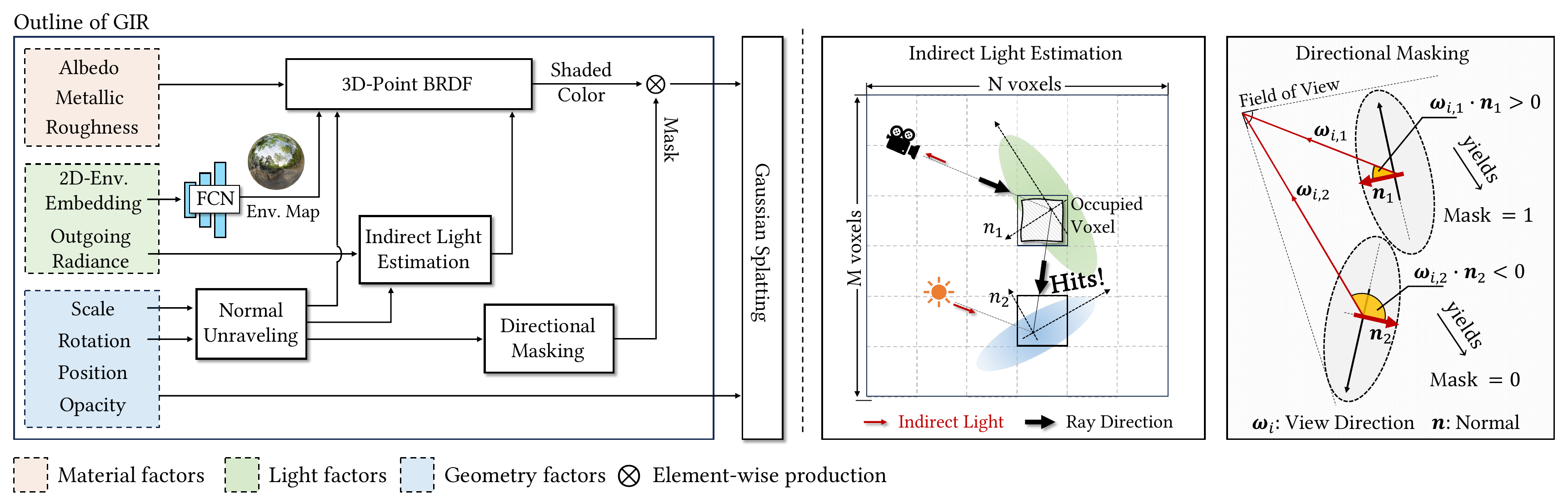}
\end{center}
\vspace{-16pt}
\caption{The outline of the proposed GIR is illustrated on the left. It optimizes three material factors, two light factors, and four geometry factors to represent a scene. An FCN is employed to learn high-quality environmental illumination by optimizing the projection of 2D-environmental embedding onto a high-resolution environmental map. {\sl Indirect Light Estimation} illustrates how a 3D Gaussian is influenced by indirect illumination. {\sl Directional Masking} illustrates calculating the mask for a 3D Gaussian's color, utilizing the unraveled normal $n$. The details are explained in Sec.~\ref{sec:method:normal_estimation}.}
\vspace{-6pt}
\label{fig: pipeline}
\end{figure*}

\section{Preliminary}
\label{sec:preliminary}
\label{sec:method:rendering}
\noindent \textbf{3D Gaussian Splatting.}
3DGS~\cite{kerbl20233d} represents a scene with a collection of 3D Gaussians, and renders a view using a differentiable splatting that projects 3D Gaussians to a 2D screen. Each 3D Gaussian $G(\vx)$ is defined by a full 3D covariance matrix $\bm{\Sigma}$ established in world space and centered at the point  $\mu$, i.e., $G(\vx) = e^{-\frac{1}{2}(\vx-\bm{\mu})^\intercal \bm{\Sigma}^{-1}(\vx-\bm{\mu})}$.
To effectively obtain $\bm{\Sigma}$ for a 3D Gaussian, 3DGS assumes that $\bm{\Sigma}$ describes the configuration of an ellipsoid corresponding to the 3D Gaussian. Then, a scaling matrix $\mS$ and a rotation matrix $\mR$ are used to parameterize $\bm{\Sigma}$ by $\bm{\Sigma} = \mR \mS\mS^\intercal \mR^\intercal$.
When rendering an image using 3D Gaussian splatting, the color $C$ for a pixel is computed by blending $N$ sorted 3D Gaussians, formally written as $C = \sum\nolimits_{i\in N} T_i \alpha_i \vc_i$,
where $T_i = \prod_{j=1}^{i-1} (1-\alpha_j)$ represents the transmission of the $i^{th}$ sorted 3D Gaussian. The variable $\vc_i$ represents the color, and $\alpha_i$ is computed based on the opacity $o_i$ by multiplying it with the covariance matrix ${\bm \Sigma_i}$ projected onto the 2D screen. Notably, 3DGS does not form a normal for any 3D Gaussian during training. 

\vspace{4pt}
\noindent \textbf{BRDF Rendering.}
In physically-based rendering, the BRDF rendering equation calculates the outgoing light from a 3D Gaussian with point $\vx$ based on the given BRDF properties. This equation establishes a relationship between the incident and outgoing light from the 3D Gaussians.
\begin{equation}
    \centering
    L_o\left(\boldsymbol{\omega}_{\boldsymbol{o}}, \vx \right)=\int_{\Omega} f\left(\boldsymbol{\omega}_{\boldsymbol{o}}, \boldsymbol{\omega}_{\boldsymbol{i}}, \vx \right) L_i\left(\boldsymbol{\omega}_i, \vx \right)\left(\boldsymbol{\omega}_i \cdot \vn \right) d \boldsymbol{\omega}_i,
    \label{eq:rendering}
\end{equation}
where the viewing direction of the outgoing light is represented by $\boldsymbol{\omega}_{\boldsymbol{o}}$, $L_i$ corresponds to the incident light coming from direction $\boldsymbol{\omega}_{\boldsymbol{i}}$, and $f$ represents the BRDF properties of the point. The integration domain is the upper hemisphere $\Omega$ defined by the point $\vx$ and its normal $\vn$. The BRDF property $f$ can be divided into diffuse and specular components according to the Disney micro-facet BRDF model~\cite{burley2012physically} as follows.
\begin{equation}
   \label{equ:disney_brdf}
   f\left(\boldsymbol{\omega}_{\boldsymbol{o}}, \boldsymbol{\omega}_{\boldsymbol{i}} \right) = (1-m) \frac{\va}{\pi} + \frac{DFG}{4(\boldsymbol{\omega}_i\cdot \vn)(\boldsymbol{\omega}_o \cdot \vn)},
\end{equation}
where $\va$ is the albedo color, $D$ is the normal distribution function, $F$ is the Fresnel term and $G$ is the geometry term. The detailed computations of $D, F, G$ are related to roughness $r$ and metallic $m$, which can be referred to~\cite{walter2007GGXmicrofacet}.

\section{Method}
\label{sec:method}
The outline of the proposed GIR is shown in Fig.~\ref{fig: pipeline}. In the following sections, we first introduce rendering and light representation in Sec.~\ref{sec:rendering_and_light}. Subsequently, we explain the details in Sec.~\ref{sec:method:normal_estimation} and~\ref{sec:method:optimization}.

\subsection{Rendering and Lighting}
\label{sec:rendering_and_light}
\noindent \textbf{Pre-integrated Rendering.}
To enable the physically-based rendering of 3D Gaussians, we parameterize each Gaussian by an additional set of terms to optimize, i.e., albedo color $\va\in [0,1]^3$, roughness $r\in [0, 1]$, and metallic $m \in [0, 1]$. 
The diffuse color is
\begin{equation}
    \vc_{\text{diffuse}} = \va (1-m) \int_{\Omega} L(\boldsymbol{\omega}_i )\frac{\boldsymbol{\omega_i} \cdot \vn}{\pi} d \boldsymbol{\omega}_i,
\label{eq:PIR-2}
\end{equation}
which can be prefiltered for accelerated computation. To simplify the computation of specular color in 3DGS, we employ the \textit{split-sum} approximation~\cite{karis2013ue4}, which has been proved effective in many industrial applications and previous works~\cite{munkberg2022extracting, liu2023nero}. This approximation allows us to separate the integral of the product of lights and BRDFs into two separate integrals, each of which can be pre-calculated and stored for efficient querying. Specifically, the specular term can be calculated by:
\begin{equation}
\begin{aligned}
\vc_{\text{specular}}  &= \int_{\Omega}  \frac{DFG}{4(\boldsymbol{\omega}_i\cdot \vn)(\boldsymbol{\omega}_o\cdot \vn)} L(\boldsymbol{\omega}_i)(\boldsymbol{\omega}_i\cdot \vn) d\boldsymbol{\omega}_i \\
 & \approx \int_{\Omega} L(\boldsymbol{\omega}_i) D(r, \vt) d\boldsymbol{\omega}_i \cdot \int_{\Omega} \frac{DFG}{4(\boldsymbol{\omega}_i\cdot \vn)} d \boldsymbol{\omega}_i,
\label{eq:PIR-splitsum}
\end{aligned}
\end{equation}
where $\vt$ is the reflective direction given the incident light direction $\boldsymbol{\omega_i}$ and normal $\vn$.
According to~\cite{karis2013ue4}, by ignoring lengthy reflections at grazing angles, we can pre-filter the environment map with importance sampling the GGX normal distribution. The environment BRDF can also be pre-integrated by substituting in Schlick’s Fresnel~\cite{schlick1994inexpensive} and saving as a 2D LUT (Look-Up-Table) that only correlates with roughness $r$ and viewing direction $\boldsymbol{\omega_o}$. Based on all the computations above, we can obtain the final shaded color $\hat{\vc} = \vc_\text{diffuse} + \vc_{\text{specular}}$.

\vspace{4pt}
\noindent \textbf{Light Representation.}
Compact low-frequency illumination models based on SH~\cite{basri2003lambertian,ramamoorthi2001efficient} have traditionally been favored in real-time rasterization methods due to their ability to represent smooth, low-frequency lighting variations efficiently. In recent years, Spherical Gaussians (SG)~\cite{wang2009all} have gained attention for their effectiveness in capturing sparse, high-frequency features and have been applied in neural inverse rendering techniques~\cite{jin2023tensoir, wu2023nefii, zhang2021physg, zhang2022invrender}. However, it is worth noting that the SG representation may face challenges when dealing with complex natural illumination conditions, as it may not scale optimally in such scenarios.
As multiple bounces exist between 3D Gaussians, to tackle the difficulty in simulating indirect illumination, we propose to model the indirect illumination using an approximate approach. This results in separately modeling the light as direct light $L_{\text{dir}}$ and indirect light $L_{\text{ind}}$. 
In particular, the incident illumination is modeled by
\begin{equation}
    L_i(\boldsymbol{\omega}_i, \vx) = V(\boldsymbol{\omega}_i, \vx)\cdot L_{\text{dir}}(\boldsymbol{\omega}_i) + L_{\text{ind}}(\boldsymbol{\omega}_i, \vx),
    \label{eq: light_representation}
\end{equation}
where $V(\boldsymbol{\omega}_i, \vx)$ represents light visibility from the outside of the bounding sphere of the scene.
The direct light contributes a significant part to the rendering of the scene. We use image-based light representation to model the direct light $L_\text{dir}$, which can be efficiently pre-integrated using mipmap stored at different roughness scales.

\subsection{3D Gaussian Inverse Rendering}
\label{sec:method:normal_estimation}
\noindent \textbf{Normal Unraveling.} 
Existing approaches often rely on continuous surface representations, such as meshes or signed distance fields (SDF), to compute normals. However, the original 3DGS approach lacks simultaneous optimization of the normal $\vn$. Estimating normals from discrete entity points is crucial for inverse rendering, but it poses a challenge since all 3D Gaussians are optimized independently.
In our observation, the maximum cross-section of an ellipsoid, oriented towards a particular angle, significantly contributes to the visible views during rendering. As a result, we propose to leverage the shortest axis of an ellipsoid to represent the point normal.

We can compute the eigenvalues and eigenvectors of the positive definite matrix $\bm{\Sigma}$ using spectral decomposition for each 3D Gaussian. Among these eigenvectors, we identify the one corresponding to the smallest eigenvalue as the point normal. We further simplify it by representing the eigenvalues as $\mS^2$. Hence, we select the shortest rotation matrix $\mR$ axis to approximate the point normal $\vn$. Notably, the smallest eigenvector and the actual normal are often not well aligned in 3DGS. This discrepancy emerges because there is no need to distinguish the sides when projecting 3D Gaussians onto 2D screen pixels. e.g., in 3DGS, two Gaussians having opposite normals both contribute to color blending by splatting. However, in PBR, the Gaussian with a back-view normal should not contribute to the shaded color. Therefore, it is crucial to explicitly define the orientation for each 3D Gaussian in conjunction with optimizing the normal $\vn$ and the rotation matrix $\mR$.
To determine the visibility of a point, we compute the dot product between the normal $\vn$ 
and the view direction $\bm\omega_o = \vo - \bm{\mu}$, where $\bm{\mu}$ represents the mean position of the 3D Gaussian and $\vo$ is the viewpoint. We can define visible and invisible points based on the value of $\vn\cdot\bm\omega_o$, and obtain a boolean visible mask by a Heaviside function $U^+$, i.e., $M=U^+(\vn\cdot\bm\omega_o)$. Refer to the illustration on the right side of Fig.~\ref{fig: pipeline} for a clearer understanding.
We can then modify the splatting function of 3DGS to incorporate the directional masking $M$:
\begin{equation}
\label{equ: normal}
    C = \sum\nolimits_{i\in N} T_i  \alpha_i (M_i\va_i + (1-M_i)\vc_\epsilon).
\end{equation}

For those Gaussians with back-view normals, we set a random color $\vc_\epsilon$ instead of black to distinguish masked Gaussians from black Gaussians. 
This implicitly enforces the surface normals to be oriented towards the correct directions by its surrounding 3D Gaussians or the newly split 3D Gaussians, regularizing to obtain sufficiently accurate normals for the scene.
Our normal unraveling process only leverages the inherent properties of 3DGS. As a result, the supervision for normal estimation is solely based on color reconstruction, making it self-supervised without requiring external supervision.

\vspace{4pt} 
\noindent \textbf{PBR Estimation.}
As explained in the preliminary section, we use roughness-metallic workflow in our PBR rendering. This requires our method to effectively estimate the PBR parameters, i.e., metallic and roughness, related to the BRDF rendering discussed in Sec.~\ref{sec:method:rendering}. As we have defined metallic $m$ and roughness $r$ as two trainable variables within each 3D Gaussian, the splatting equation can be rewritten as follows:
\begin{equation}
    C = \sum\nolimits_{i\in N} T_i M_i \alpha_i ((\vc_i*2^\beta)^\gamma + (1-M_i)\vc_\epsilon).
\end{equation}

We replace the albedo color $\va$ with the shaded color $\vc$ after applying the 3D-Point BRDF to each 3D Gaussian. Additionally, we convert the linear color space to the sRGB color space for the rendered image, using a gamma value ($\gamma$) of 2.4 and an exposure value ($\beta$), with a default of 0.

\vspace{4pt}  
\noindent \textbf{Direct Illumination Estimation.} 
We employ a pre-integrated rendering technique that utilizes mipmap cube maps to represent direct illumination. This involves using a high-resolution HDR image (e.g., $1024\times 2048$) and pre-integrating it at different mipmap levels based on roughness. However, training directly on the high-resolution HDR environmental map often leads to local optima due to the large number of trainable parameters and the unstable gradients propagated through the numerous trainable 3D Gaussians.

To address this challenge, we simplify the training process by incorporating an additional compact neural network to bridge the desired high-resolution HDR and training entity, similar to the representation used in image restoration~\cite{ulyanov2018deep}. Specifically, we introduce a differentiable function $g_\theta: \mathbb{R}^{128\times64\times128}\rightarrow\mathbb{R}^{3\times1024\times2048}$ that generates and optimizes environment light maps $l$ for each scene. These environment light maps are then converted into cube maps and utilized within the physically-based rendering pipeline, taking advantage of the mipmaps. In our implementation, $g_\theta$ begins with a learnable constant and is implemented as a Fully Convolutional Network (FCN) with up-sampling layers. The FCN comprises convolution layers with LeakyReLU activation functions, and after two convolution layers, the feature map resolution is doubled, resulting in high-fidelity and smooth environment light maps.

\vspace{4pt} 
\noindent \textbf{Indirect Illumination Estimation.} 
Direct illumination refers to light coming from distant sources, while indirect illumination arises from self-occlusion within objects. Modeling indirect illumination accurately within the context of 3DGS presents a significant challenge. We propose a simple yet effective method to approximate indirect illumination among 3D Gaussians to address this.
Traditionally, recovering indirect illumination involves recursive path tracing on surfaces, which is computationally demanding. Our approach introduces an approximate ray tracing technique to simulate spatially varying indirect illumination originating from 3D Gaussians. Our key insight is to decompose indirect illumination into two components: SH coefficients represented as light and a view-dependent occlusion mask that determines the visibility of indirect illumination. 

To compute the indirect incoming light $L_{\text{ind}}(\bm\omega_i,\vx)$ at position $\vx$ from view direction $\bm\omega_o$, we sample rays over the hemisphere by ray tracing. However, as the number of bounces increases, the computational complexity of tracing and rendering exponentially grows with the number of samples. To address this challenge, we propose to leverage trainable SH coefficients to encode the view direction $\bm\omega_o$, enabling the simulation of integrated colors at point $\vx$. The configuration of SH coefficients is identical to that employed in 3DGS. This approach enables us to approximate indirect illumination while reducing the computational complexity.
Per the light representation (Equ.~\ref{eq: light_representation}), we employ an occlusion mask $M_{\text{occ}}$ to determine whether indirect lights should contribute to the rendering process. 

As illustrated in Indirect Light Estimation of Fig.\ref{fig: pipeline}, the occlusion mask is computed through a one-step ray tracing procedure. Before starting the ray tracing process, we fill a binary voxel grid to simulate geometry based on the shape of 3D Gaussians. We generate the voxel grid by computing a bounding box for each 3D Gaussian using its covariance matrix $\bm\Sigma\times3$, according to the Pauta criterion~\cite{peircecriterion}, which detects outliers with a confidence probability of $99.7\%$. We then consider all voxels within this bounding box as representing the 3D Gaussian. Subsequently, we transform the 3D Gaussians into binary grids. The resolution of the voxel grid is $128^3$, and its scale adjusts to fit the geometry of the 3D scenes. When a ray is traced from the point $\vx$ towards the reflection direction $\omega_i = (\vn\cdot\bm\omega_o)\vn-\bm\omega_o$ and checked for occlusion by voxel grid within a radius of the bounding sphere. If the ray encounters another voxel containing 3D Gaussians, the occupied voxel of point $\vx$ in the view direction $\omega_o$ is flagged as occluded. Specifically, we sample $64$ points from point $\vx$ along the reflection direction $\omega_i$, and a sample point $\hat{x} = \vx + t\omega_i$ falls within a filled voxel. We sample 128 rays within a hemisphere for diffuse lighting to calculate occlusion for shadow computation. We compute an occlusion mask $M_{\text{occ}}$ for each step along the viewing direction $\omega_o$ for specular lighting. If the specular light encounters an obstacle as indicated by the $M_{\text{occ}}$, we utilize trainable SHs that encode the viewing direction $\bm\omega_o$ to simulate indirect illumination.

\subsection{Regularization and Optimization}
\label{sec:method:optimization}
\noindent {\bf Disparity Smoothness Regularization.}
To achieve smooth normal predictions, we incorporate an edge-aware smoothness term as described in~\cite{zhan2018unsupervised}. This encourages local smoothness by applying an $L1$ penalty on the disparity gradients $\partial\vn$, as formulated below:
\begin{equation}
\mathcal{L}_\text{s}(\vn) = \frac{1}{HW} \sum\nolimits_{i \in H, j \in W} \left | \partial_x \vn_{ij} \right | e^{-\left | \partial_x I_{ij} \right | } + \left | \partial_y \vn_{ij} \right | e^{-\left | \partial_y I_{ij} \right | },
\end{equation}
where $I$ denotes the ground truth image, and $\partial_x(.)$ and $\partial_y(.)$ are the horizontal and vertical gradients, respectively. Similarly, we apply this smoothness term to the albedo color $\mathcal{L}_\text{s}(\va)$, metallic $\mathcal{L}_\text{s}(m)$, and roughness $\mathcal{L}_\text{s}(r)$. These terms ensure consistency in the independent optimization of each Gaussian component.

% NVS
\begin{figure*}[t]
\begin{center}
\includegraphics[width=0.9\linewidth]{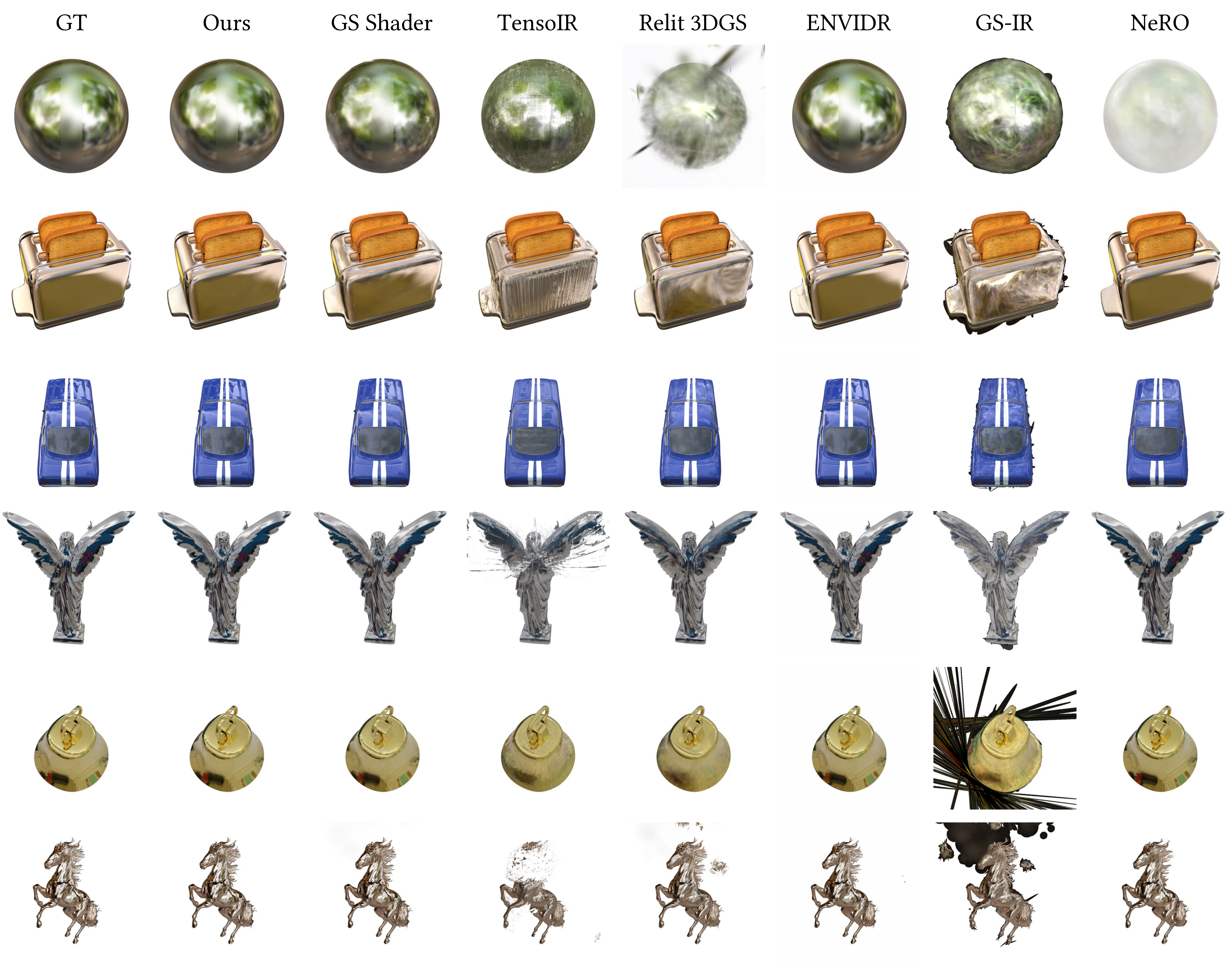}
\end{center}
\vspace{-20pt}
   \caption{Qualitative comparison of NVS on the Shiny Blender (first three rows) and Glossy Blender (last three rows) datasets.}
\vspace{-10pt}
\label{fig:NVS_refnerf_nero}
\end{figure*}

\vspace{4pt} 
\noindent {\bf Environment Light Regularization.}
Besides regularizing properties on the target scene, we also impose regularization on direct illumination. The light regularization term $\mathcal{L}_l$ is specifically devised to enforce neutral and white lighting in the scene~\cite{munkberg2022extracting}. To achieve this, we calculate the mean of the cube maps representing the environment light and minimize the discrepancy between the RGB values and their mean values. The equation used for this purpose is as follows:
\begin{equation}
    \mathcal{L}_l = \frac{1}{H^lW^l}\sum\nolimits_{i\in H^l,j\in W^l} |l_{i,j}-\bar {l}_{i,j}|,
\end{equation}
where ${\bar l}_{i,j}$ is calculated by averaging RGB channels for each pixel on the environment light map $l$.

\vspace{4pt} 
\noindent {\bf Image Reconstruction Loss.}
In terms of optimization, we utilize a combined loss function for image reconstruction, which incorporates both Mean Absolute Error (MAE) represented as $\mathcal{L}_{\text{MAE}}$ and a D-SSIM term denoted as $\mathcal{L}_{\text{SSIM}}$. This loss function comprehensively evaluates the image reconstruction process by considering the absolute differences and structural similarities between the generated image and the ground truth. By incorporating these components, we aim to enhance the fidelity and perceptual quality of the reconstructed image.

\definecolor{tabfirst}{rgb}{1, 0.7, 0.7} % red
\definecolor{tabsecond}{rgb}{1, 0.85, 0.7} % orange
\definecolor{tabthird}{rgb}{1, 1, 0.7} % yellow

\begin{table}[t]
\centering
{
%\fontsize{6.5pt}{7.8pt}\selectfont
\caption{The evaluation results of novel view synthesis (\textbf{PSNR}$\uparrow$) tested on Shiny Blender~\cite{verbin2022ref} dataset.}
\vspace{-6pt}
\label{tab:refnerf}
\resizebox{\linewidth}{!}{
\begin{tabular}{l|cccccc|c}
\toprule
Method   & Car   & Ball  & Helmet & Teapot & Toaster & Coffee & Avg.  \\
\midrule
NVDiffRec~\cite{munkberg2022extracting} & 24.63 & 18.04 & 27.12  & 40.13  & 23.79 & 30.77  & 27.41 \\
NVDiffMC~\cite{hasselgren2022nvdiffrecmc}  & 25.14 & 17.35 & 25.77 & 37.45  & 21.82 & 29.01 & 26.09 \\
TensoIR~\cite{jin2023tensoir} & 26.04 & 21.89 & 25.17 & 43.01 & 18.31 & \cellcolor{tabthird}31.71 & 27.69 \\
EnvIDR~\cite{liang2023envidr}    & \cellcolor{tabthird}28.39 & \cellcolor{tabfirst}41.71 & \cellcolor{tabfirst}32.62  & \cellcolor{tabthird}43.57  & \cellcolor{tabthird}24.51   & 29.11  & \cellcolor{tabsecond}33.32 \\
NeRO~\cite{liu2023nero}      & 25.24 & \cellcolor{tabthird}32.80 &\cellcolor{tabthird}28.35  & 42.13  & \cellcolor{tabsecond}25.19   & 31.24  & \cellcolor{tabthird}30.83 \\
GS Shader~\cite{jiang2023gaussianshader} & \cellcolor{tabsecond}28.45 & 29.31 & 28.33 & 43.42 & 23.02 & 31.45 & 30.66 \\
GS-IR~\cite{liang2023gs} & 25.57 & 19.41 & 25.06 & 38.57 & 18.83 & 30.66 & 26.35 \\
Relit 3DGS~\cite{gao2023relightable} & 26.55 & 20.18 & 26.92 & \cellcolor{tabsecond}43.59 & 19.91 & \cellcolor{tabsecond}31.92 & 28.18 \\
\midrule
Ours      &  \cellcolor{tabfirst}30.78 & \cellcolor{tabsecond}37.13 & \cellcolor{tabsecond}31.06  & \cellcolor{tabfirst}44.81  & \cellcolor{tabfirst}27.44   & \cellcolor{tabfirst}35.81  & \cellcolor{tabfirst}34.51 \\
\bottomrule
\end{tabular}
}
}
\vspace{-6pt}
\end{table}

\section{Experiments}
\label{sec:experiments}
\subsection{Implementation and Baselines}
We refer to the implementation details in the supplementary materials. To ensure reproducibility, we will publicly release the source code at \href{https://github.com/guduxiaolang/GIR}{https://github.com/guduxiaolang/GIR}. We utilize NeRFactor~\cite{zhang2021nerfactor}, NVDiffRec~\cite{munkberg2022extracting}, NVDiffrecMC~\cite{hasselgren2022nvdiffrecmc}, InvRender~\cite{zhang2022invrender}, ENVIDR~\cite{liang2023envidr}, NeRO~\cite{liu2023nero}, TensoIR~\cite{jin2023tensoir}, Gaussian Shader~\cite{jiang2023gaussianshader}, GS-IR~\cite{liang2023gs}, and Relightable 3D Gaussian~\cite{gao2023relightable} as 
our baseline methods, taking into account their popularity and the code availability.

\begin{table}[t]
\centering
{
%\fontsize{6.5pt}{7.8pt}\selectfont
\caption{The evaluation results of novel view synthesis (\textbf{PSNR}$\uparrow$) tested on Glossy Blender dataset~\cite{liu2023nero} dataset.}
\vspace{-6pt}
\label{tab:invrender}
\resizebox{\linewidth}{!}{
\begin{tabular}{l|cccccccc|c}
\toprule
Method & Angel & Bell & Cat & Horse & Luyu & Potion & Tbell & Teapot & Avg. \\
\midrule
NVDiffRec~\cite{munkberg2022extracting} & 27.70 & 30.69 & 30.22 & 27.07 & 27.31 & 29.43 & 21.33 & 24.55 & 27.29 \\
NVDiffMC~\cite{hasselgren2022nvdiffrecmc} & 26.25 & 27.59 & 28.59 & 25.69 & 26.84 & 27.99 & 25.90 & 23.42 & 26.53 \\
TensoIR~\cite{jin2023tensoir} & 19.11 & 24.27 & 28.43 & 16.78 & 25.86 & 28.76 & 22.94 & 20.97 & 23.39 \\
EnvIDR~\cite{liang2023envidr} & \cellcolor{tabsecond}29.26 & \cellcolor{tabfirst}32.56 & 32.09 & \cellcolor{tabsecond}28.62 & \cellcolor{tabsecond}29.61 & \cellcolor{tabsecond}33.21 & \cellcolor{tabsecond}29.79 & \cellcolor{tabsecond}28.32 & \cellcolor{tabsecond}30.42 \\
NeRO~\cite{liu2023nero} & \cellcolor{tabthird}29.13 & \cellcolor{tabthird}31.40 & \cellcolor{tabsecond}32.83 & 28.33 & 28.40 & \cellcolor{tabfirst}33.32 & \cellcolor{tabthird}28.10 & \cellcolor{tabfirst}28.37 & \cellcolor{tabthird}29.99 \\
GS Shader~\cite{jiang2023gaussianshader} & 27.75 & 30.72 & \cellcolor{tabthird}32.60 & \cellcolor{tabthird}28.57 & \cellcolor{tabthird}28.52 & 30.29 & 26.81 & 25.13 & 28.80 \\
GS-IR~\cite{liang2023gs} & 23.08 & 24.22 & 28.12 & 23.85 & 25.40 & 27.76 & 22.28 & 21.08 & 24.47 \\
Relit 3DGS~\cite{gao2023relightable} & 24.04 & 25.45 & 25.74 & 25.12 & 26.79 & 29.76 & 23.89 & 24.45 & 25.65 \\
\midrule
Ours & \cellcolor{tabfirst}30.00 & \cellcolor{tabsecond}31.61 & \cellcolor{tabfirst}34.26 & \cellcolor{tabfirst}29.77 & \cellcolor{tabfirst}29.71 & \cellcolor{tabthird}32.92 & \cellcolor{tabfirst}30.01 & \cellcolor{tabthird}25.99 & \cellcolor{tabfirst}30.53 \\
\bottomrule
\end{tabular}
}
}
\vspace{-6pt}
\end{table}

\subsection{Metrics and Datasets}
\noindent {\bf Metrics.}
We assess the inverse rendering capabilities by evaluating the performance in novel view synthesis, scene relighting, and albedo estimation tasks. To measure the quality of these tasks, we utilize Peak signal-to-noise ratio (PSNR), Structural Similarity (SSIM), and Learned Perceptual Image Patch Similarity (LPIPS).

% relight
\begin{figure*}[t]
\begin{center}
\includegraphics[width=0.9\linewidth]{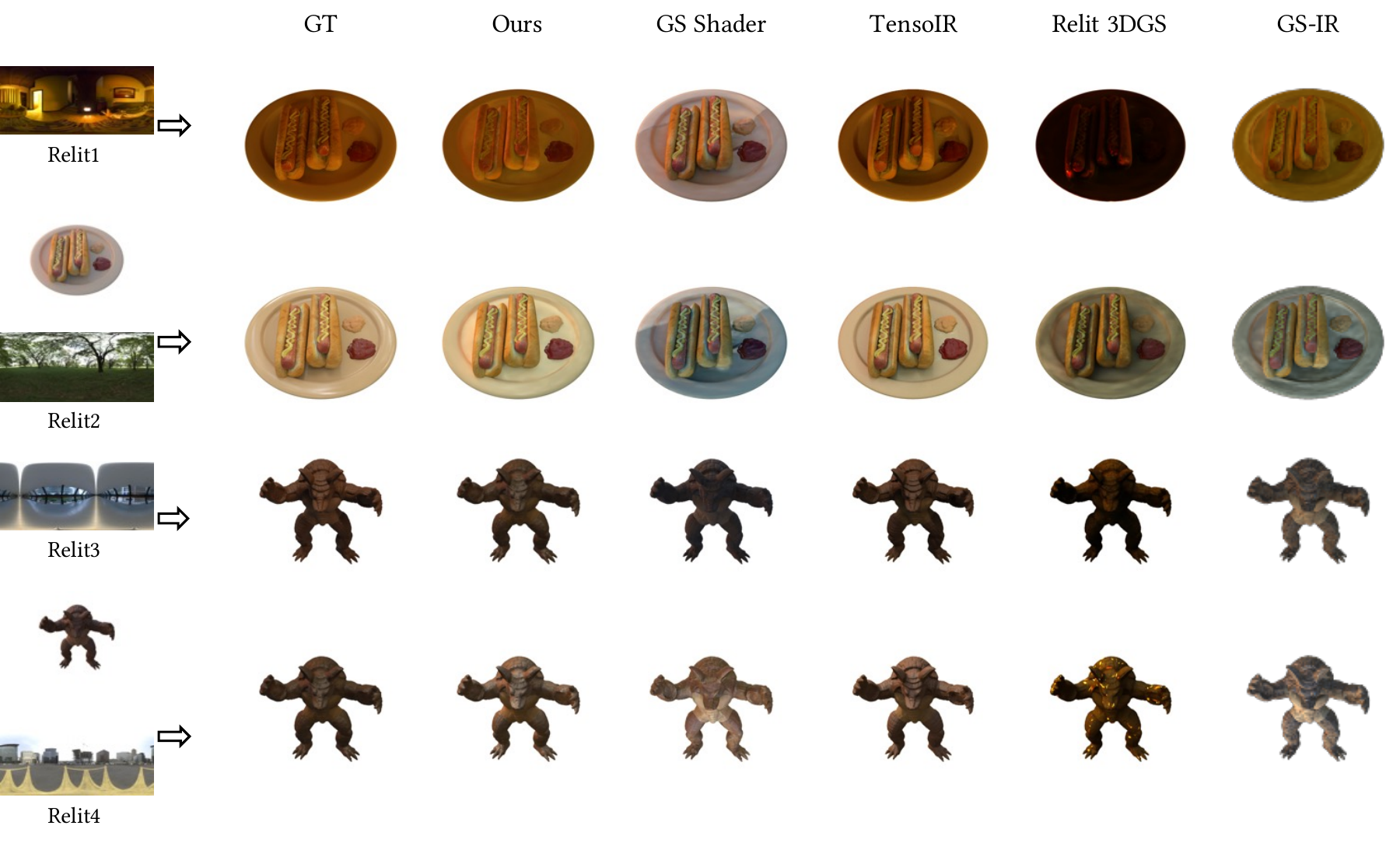}
\end{center}
    \vspace{-16pt}
   \caption{Qualitative comparison of relighting on the TensoIR dataset. The first column shows four new environmental lights and two reference images.}
\label{fig:relighting_tensoir}
\vspace{-10pt}
\end{figure*}

\vspace{4pt} 
\noindent {\bf Datasets.}
We evaluate our approach and all baselines using three synthetic and one real datasets: 1) the TensoIR dataset~\cite{jin2023tensoir}, 2) the Shiny Blender dataset~\cite{verbin2022ref}, 3) the Glossy Blender dataset~\cite{liu2023nero}, 4) the Objects with Lighting (OWL) dataset~\cite{Ummenhofer2024OWL}. TensoIR and Shiny Blender datasets comprise 100 training images and 200 testing images per object. The Glossy Blender dataset comprises 128 training images and 8 testing images per object, and the OWL dataset includes 40-60 training images and 9 testing images under various lighting conditions per object. We strictly adhere to the predefined training and testing splits provided for these datasets. These datasets encompass four distinct types: dielectric, conductor, a combination of dielectric and conductor, and real data.

\begin{table}[t]
\centering
{
%\fontsize{6.5pt}{7.8pt}\selectfont
\caption{Quantitative comparisons on TensoIR dataset~\cite{jin2023tensoir}.}
\vspace{-6pt}
\label{tab:tensoir}
\resizebox{\linewidth}{!}{
\begin{tabular}{l|ccc|ccc}
\toprule
\multirow{2}{*}{Method} & \multicolumn{3}{c|}{Novel View Synthesis} & \multicolumn{3}{c}{Relighting}          \\
                        &  PSNR$\uparrow$   & SSIM$\uparrow$  & LPIPS$\downarrow$   & PSNR$\uparrow$ & SSIM$\uparrow$ & LPIPS$\downarrow$ \\ 
\midrule

NeRFactor~\cite{zhang2021nerfactor} & 24.679 & 0.929 & 0.121 &  
23.383 & 0.908 & 0.131 \\
InvRender~\cite{zhang2022invrender} & 27.368 & 0.934 & 0.089 &  \cellcolor{tabthird}23.973 & 0.901 & 0.101\\
NVDiffrec~\cite{munkberg2022extracting} & 30.019 & 0.962 & 0.052 & 19.881 & 0.879 & 0.092 \\
TensoIR~\cite{jin2023tensoir} & 35.143 & \cellcolor{tabthird}0.976 & \cellcolor{tabthird}0.040 & \cellcolor{tabsecond}28.586 & \cellcolor{tabfirst}0.945 & \cellcolor{tabsecond}0.080 \\
EnvIDR\tablefootnote{EnvIDR does not converge in our experiments.} ~\cite{liang2023envidr}& N/A & N/A & N/A & N/A & N/A & N/A \\
NeRO~\cite{liu2023nero}& 32.602 & 0.933 & 0.082 & 21.662 & 0.836 & 0.151 \\
GS Shader~\cite{jiang2023gaussianshader} & \cellcolor{tabsecond}37.573 & \cellcolor{tabfirst}0.984 & \cellcolor{tabfirst}0.022 & 23.372 & \cellcolor{tabthird}0.910 & \cellcolor{tabthird}0.084 \\
GS-IR~\cite{liang2023gs} & 34.973 & 0.962 & 0.043 & 23.881 & 0.873 & 0.109 \\
Relit 3DGS~\cite{gao2023relightable} & \cellcolor{tabthird}37.131 & \cellcolor{tabfirst}0.984 & \cellcolor{tabfirst}0.022 & 21.404 & 0.860 & 0.127 \\
\midrule
Ours & \cellcolor{tabfirst}37.626 & \cellcolor{tabsecond}0.980 & \cellcolor{tabsecond}0.026 & \cellcolor{tabfirst}28.883 & \cellcolor{tabsecond}0.943 & \cellcolor{tabfirst}0.059\\ 
\bottomrule
\end{tabular}}
}
\begin{flushleft}
Our method exhibits effectiveness in evaluating novel view synthesis and relighting tasks. For a fair comparison, all novel view synthesis results are generated with physically-based rendering. The results show that our method can intrinsically factorize scenes for high-quality rendering.
\end{flushleft}
\vspace{-6pt}
\end{table}

\subsection{Evaluation on Synthetic Data}
\noindent \textbf{Novel View Synthesis (NVS).}
We showcase the performance of our proposed GIR through the evaluation of novel view synthesis (NVS). The quantitative comparisons of results are presented in Tab.\ref{tab:refnerf}, Tab.\ref{tab:invrender}, and Tab.\ref{tab:tensoir}, demonstrating the overall improvement achieved by GIR compared to baseline approaches in three datasets. Furthermore, we provide qualitative comparisons in Fig.~\ref{fig:NVS_refnerf_nero}, illustrating the high-quality scene factorization accomplished by our proposed method. We also show the estimated environment light in Fig.~\ref{fig:HDR}. We provide more novel view synthesis and estimated environment results in the supplementary materials.

\begin{table}[t]
\centering
{
%\fontsize{6.5pt}{7.8pt}\selectfont
\caption{The evaluation of relit results tested on OWL~\cite{Ummenhofer2024OWL} dataset.}
\vspace{-6pt}
\label{tab:realdataset}
\resizebox{\linewidth}{!}{
\begin{tabular}{l|ccc|ccc}
\toprule
\multirow{2}{*}{Method} & \multicolumn{3}{c|}{Same Environment} & \multicolumn{3}{c}{New Environment} \\
                        &  PSNR$\uparrow$   & SSIM$\uparrow$  & LPIPS$\downarrow$    & PSNR$\uparrow$ & SSIM$\uparrow$ & LPIPS$\downarrow$ \\ 
\midrule

NVDiffrec\cite{munkberg2022extracting} & \cellcolor{tabsecond}26.138 & \cellcolor{tabsecond}0.959 & \cellcolor{tabsecond}0.054 & \cellcolor{tabsecond}24.778 & \cellcolor{tabthird}0.948 & \cellcolor{tabsecond}0.061 \\
NVDiffMC~\cite{hasselgren2022nvdiffrecmc} & 21.592 & 0.874 & 0.152 & 20.158 & 0.844 & \cellcolor{tabthird}0.125 \\
TensoIR\cite{jin2023tensoir} & \cellcolor{tabthird}23.256 & \cellcolor{tabthird}0.953 & \cellcolor{tabthird}0.058 & \cellcolor{tabthird}23.505 & \cellcolor{tabsecond}0.951 & \cellcolor{tabsecond}0.061 \\
EnvIDR$^1$~\cite{liang2023envidr} & N/A & N/A & N/A & N/A & N/A & N/A \\
GS Shader~\cite{jiang2023gaussianshader} & 14.622 & 0.830 & 0.155 & 11.297 & 0.792 & 0.178 \\
GS-IR~\cite{liang2023gs} & 17.343 & 0.878 & 0.176 & 14.685 & 0.835 & 0.167 \\
Relit 3DGS~\cite{gao2023relightable} & 14.371 & 0.838 & 0.157 & 13.264 & 0.801 & 0.171 \\
Ours & \cellcolor{tabfirst}27.779 & \cellcolor{tabfirst}0.973 & \cellcolor{tabfirst}0.043 & \cellcolor{tabfirst}27.176 & \cellcolor{tabfirst}0.966 & \cellcolor{tabfirst}0.049 \\
\bottomrule
\end{tabular}}
}
\vspace{-6pt}
\end{table}

\vspace{4pt} 
\noindent \textbf{Relighting.} We evaluate the proposed GIR by conducting scene-relighting experiments using the test splits of all datasets. The quantitative comparison is conducted on the TensorIR dataset using five provided HDR environmental maps. The results, as shown in Tab.~\ref{tab:tensoir}, demonstrate that the proposed GIR outperforms TensoIR in terms of PSNR and LPIPS metrics while achieving a comparable performance in terms of SSIM. The qualitative results, depicted in Fig~\ref{fig:relighting_tensoir}, demonstrate the GIR's capability to produce realistic relighting outcomes. We provide more relighting results across various datasets and environmental lighting combinations in the supplementary materials.

\vspace{4pt} 
\noindent \textbf{Albedo Estimation.}
To assess the performance of the proposed GIR compared to baseline methods, we compare the estimated albedo images obtained by these approaches. The quantitative experiments are conducted on the TensoIR dataset, with the results presented in the middle section of Tab.1 in the supplementary materials. The results demonstrate that the proposed GIR performs favorably compared to the baseline methods regarding scene factorization. To comprehensively represent the albedo estimation performance, we present visualizations from the TensoIR dataset and the Shiny Blender dataset in  Fig.~\ref{fig:albedo_tensoir_refnerf}. We also show our albedo estimation results on all datasets in the supplementary materials.

% albedo
\begin{figure*}[t]
\begin{center}
\includegraphics[width=0.95\linewidth]{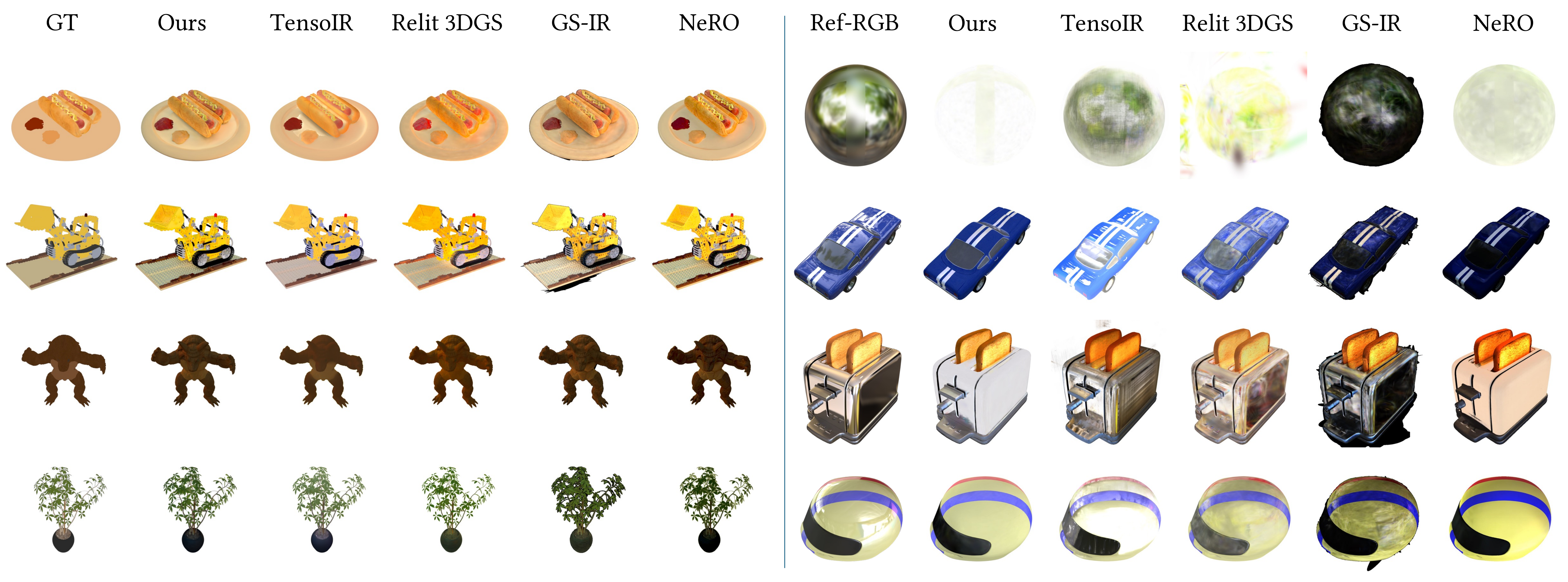}
\end{center}
\vspace{-16pt}
   \caption{Qualitative comparison of albedo. Left: TensoIR dataset. Right: Shiny Blender dataset. Since the TensoIR dataset does not provide GT albedo, we use GT images for reference. Note that GS Shader is not included in the comparison due to the absence of albedo.}
\vspace{-10pt}
\label{fig:albedo_tensoir_refnerf}
\end{figure*}

\begin{table}
\centering
{
%\fontsize{6.5pt}{7.8pt}\selectfont
\caption{The ablation study on evaluating the effectiveness of each proposed module.}
\vspace{-6pt}
\label{tab:ablation}
\resizebox{\linewidth}{!}{
\begin{tabular}{cccc|lll}
\toprule
FCN   & Geometry Reg.  & Material Reg. & Indirect Ill. & PSNR$\uparrow$  & SSIM$\uparrow$ & LPIPS$\downarrow$\\
\midrule
- & - & - & - & 32.283 & 0.958 & 0.064\\
 \ding{51} & - & - & - & 33.241 & 0.964 & 0.062\\
 \ding{51} & \ding{51} & - & - & \cellcolor{tabthird}33.432 & 0.964 & 0.061 \\
 \ding{51} & - & \ding{51} & - & 33.377 & \cellcolor{tabthird}0.965 & \cellcolor{tabthird}0.060\\
 \ding{51} & \ding{51} & \ding{51} & - & \cellcolor{tabsecond}33.759 & \cellcolor{tabsecond}0.966 & \cellcolor{tabsecond}0.059 \\
 \hline
 \ding{51} & \ding{51} & \ding{51} & \ding{51} & \cellcolor{tabfirst}34.513 & \cellcolor{tabfirst}0.971 & \cellcolor{tabfirst}0.057 \\
\midrule 
\end{tabular}
}
\begin{flushleft}
Geometry Reg. denotes regularizing depth and normal, while Material Reg. denotes regularizing roughness and metallic.
\end{flushleft}
\vspace{-6pt}
}
\end{table}

\subsection{Evaluation on Real-world Data}
\begin{figure*}[t]
\begin{center}
\includegraphics[width=0.9\linewidth]{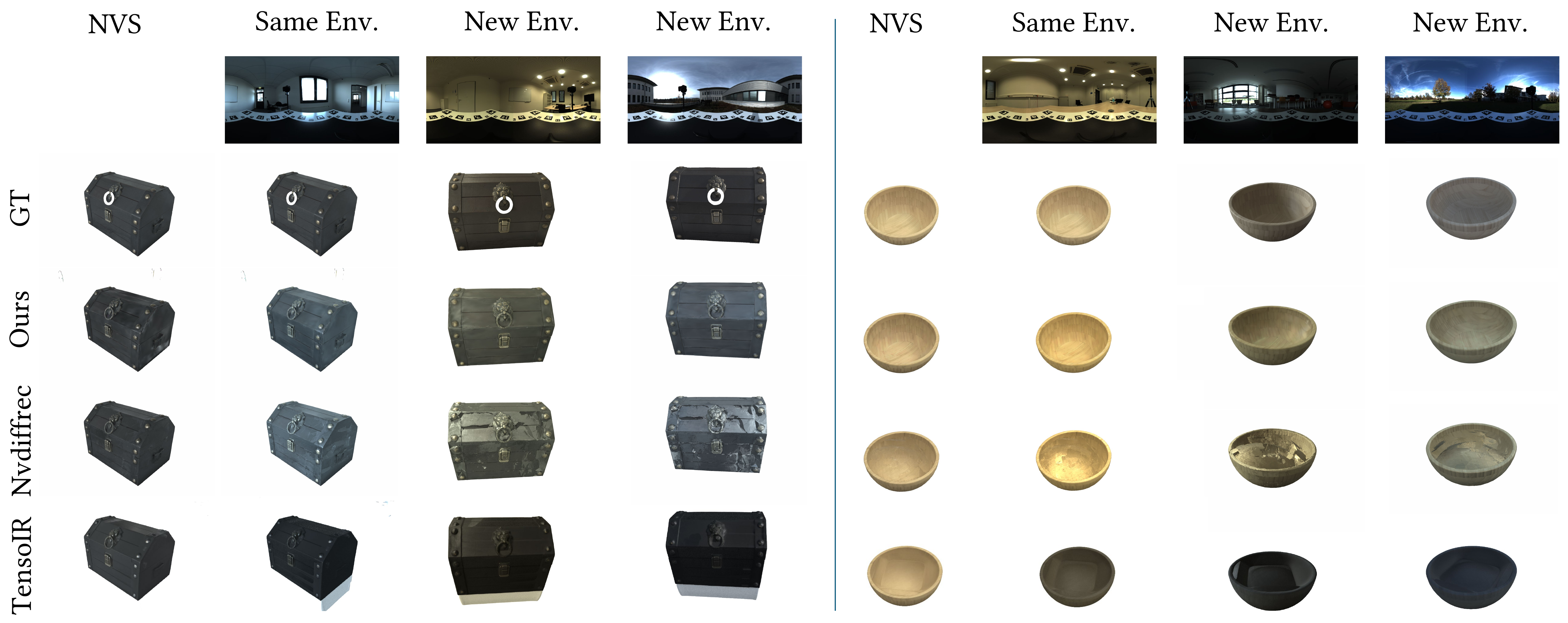}
\end{center}
    \vspace{-6pt}
   \caption{Qualitative comparison of NVS and relighting on the real-world OWL dataset. Our method demonstrates higher visual quality.}
\label{fig:real_nvs_relit}
\vspace{-10pt}
\end{figure*}

To showcase the efficacy of the proposed GIR in real-world scenarios, we conduct experiments on a real dataset, i.e., OWL~\cite{Ummenhofer2024OWL}. During testing, one of the environmental maps corresponds to the reconstructed environment but remains inaccessible during the reconstruction process. Those results are reported in the ``Same Environment'' column of Tab.~\ref{tab:realdataset}. Additionally, the dataset consists of test images captured under two novel environmental maps. Those results are presented in the ``New Environment'' column. The findings indicate that our GIR outperforms the performance of baseline methods when applied to real-world data, as shown in Fig.~\ref{fig:real_nvs_relit}.

\begin{figure}[t]
\begin{center}
%\fbox{\rule{0pt}{2.2in} \rule{0.9\linewidth}{0pt}}
   \includegraphics[width=0.8\linewidth]{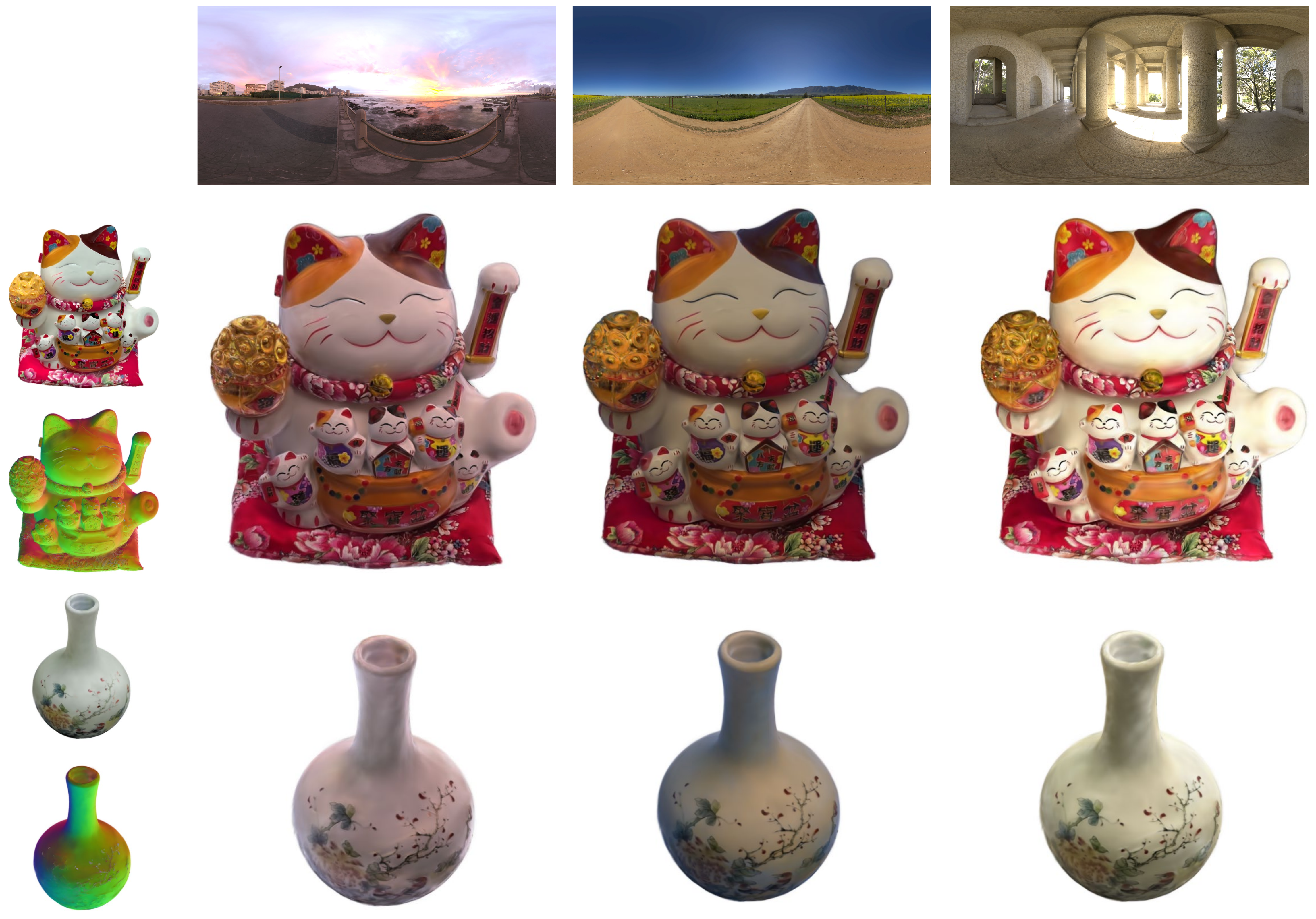}
\end{center}
\vspace{-12pt}
   \caption{The relit results produced by the proposed GIR on the real-world data captured using iPhone13 Pro Max using around 100 observed images.}
\label{fig:lucky_cat}
\vspace{-6pt}
\end{figure}

We also performed relightable scene factorization on the ``Lucky Cat'' and ``Porcelain Bottle'' datasets, captured by the authors using an iPhone 13 Pro Max. We used COLMAP~\cite{schoenberger2016sfm} to compute the camera's intrinsic and extrinsic parameters. The proposed GIR is trained using 100 posed images and their corresponding masks. We evaluate its factorization performance by relighting with three different environmental maps from a novel viewpoint. The results, shown in Fig.~\ref{fig:lucky_cat}, demonstrate the effectiveness of the proposed GIR in real-world scenarios. We offer results showcasing the relighting of the scene under a point light source in the accompanying video.

\subsection{Ablation Study}
\label{sec:ablation}
\noindent 
\textbf{Effectiveness of Each Design} 
We have proposed multiple designs in this study, which are: 1) learning to predict a high-quality environmental map using FCN; 2) a geometry regularization; 3) a material regularization; 4) an effective indirect illumination estimation, and 5) the directional masking scheme. This ablation study investigates the contribution of each design to the proposed GIR framework. To this end, GIR is trained with different configurations according to Tab.~\ref{tab:ablation} on the Shiny Blender dataset. The results are shown in Tab.~\ref{tab:ablation}. Firstly, we observe that utilizing a Fully Convolutional Network (FCN) proves to be more effective than directly optimizing the high-resolution environmental map. This is attributed to the FCN's ability to super-resolve the 2D environmental embedding, resulting in a smoother and more accurate environmental map. We show the evidence in the supplementary materials. Next, we observe that the proposed geometry and material regularization improve performance. Simultaneously, employing both yields the maximum performance gain. Furthermore, the effectiveness of the proposed indirect light estimation is validated as it improves the performance, confirming its positive impact on the results. Notably, the proposed directional masking scheme is enabled in all the experiments above since it is crucial for normal estimation. Otherwise, the overall performance degrades.

% real data

\begin{figure}[t]
\begin{center}
\includegraphics[width=0.9\linewidth]{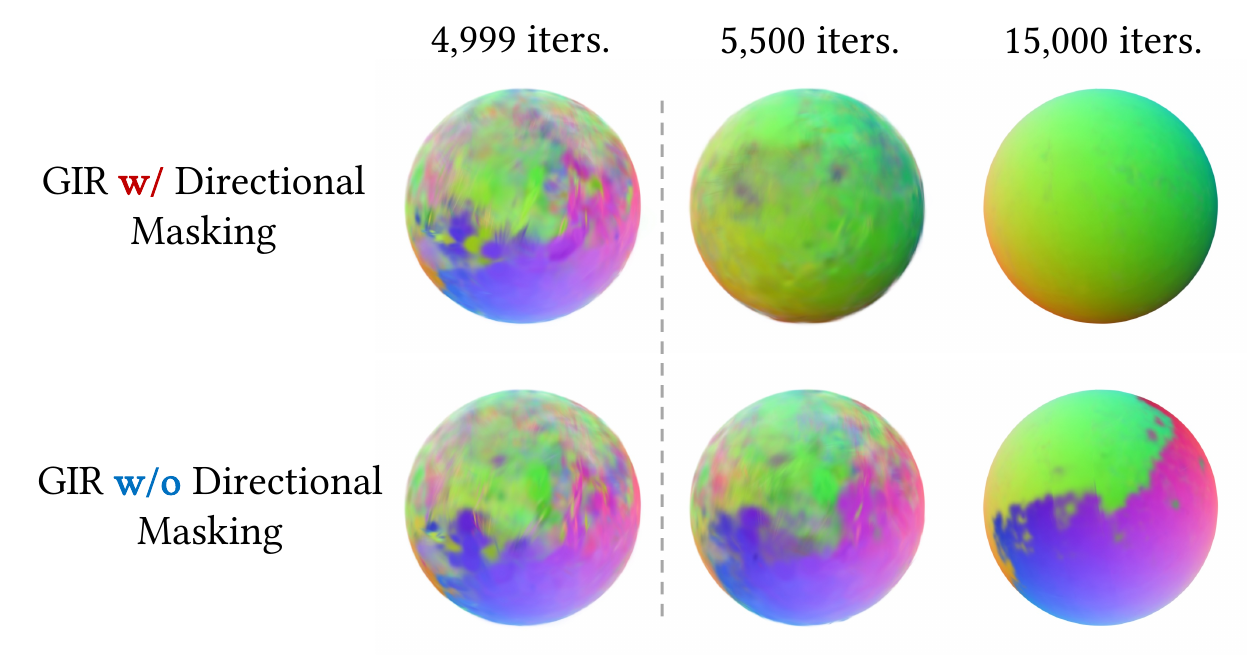}
\vspace{-16pt}
\end{center}
\caption{The results are produced w/ and w/o the proposed directional masking scheme. The comparison demonstrates the effectiveness of this scheme. A detailed explanation is provided in Sec.~\ref{sec:ablation}.}
\vspace{-6pt}
\label{fig:dir_mask}
\end{figure}

\vspace{4pt} 
\noindent \textbf{Rendering Speed.}
We thoroughly evaluate the rendering speed on a scene containing 350,000 3D Gaussians. First, we test the rendering speed in an inference environment on an NVIDIA Tesla V100 GPU. Our proposed method achieves an average rendering framerate of $92.58$ frames per second (FPS) for this scene. The original 3DGS framework demonstrates a rendering framerate of $120.13$ FPS on the same scene. Second, we developed a plugin in Unreal Engine 5 (UE5) using its Niagara system and adapted its PBR rendering equation to our metallic-roughness workflow. In this setting, the rendering speed is $78.75$ (ours) \textit{v.s.} $85.57$ (3DGS) in FPS on an NVIDIA GTX 3090 GPU, exhibiting the real-time rendering capability of our proposed inverse rendering method.

\begin{figure}[t]
\begin{center}
\includegraphics[width=0.85\linewidth]{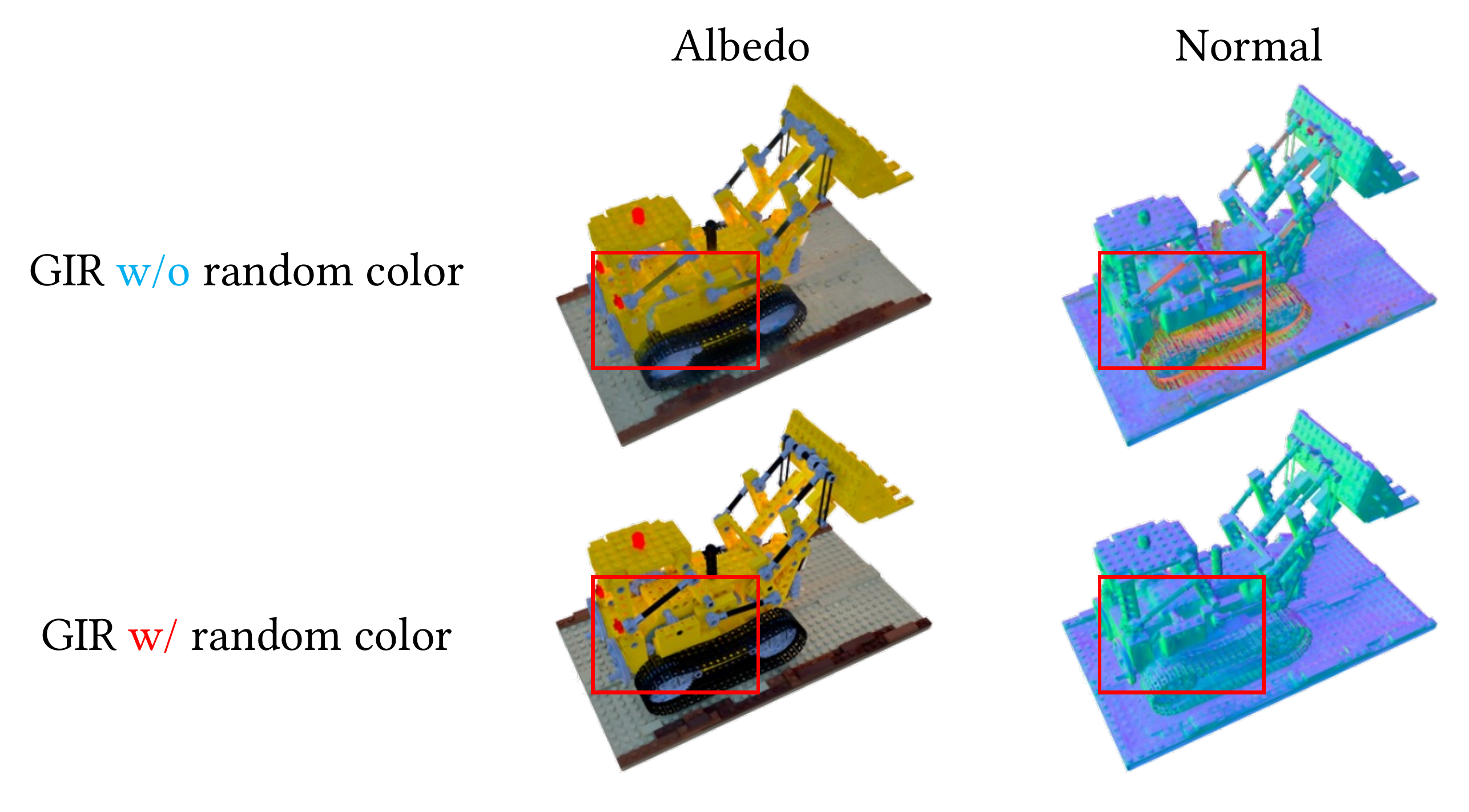}
\end{center}
\vspace{-16pt}
\caption{Visualization of the effectiveness of using random color for those Gaussians with back-view normals.}
\vspace{-10pt}
\label{fig:maks_random_color}
\end{figure}

\begin{figure}[t]
\begin{center}
\includegraphics[width=1.0\linewidth]{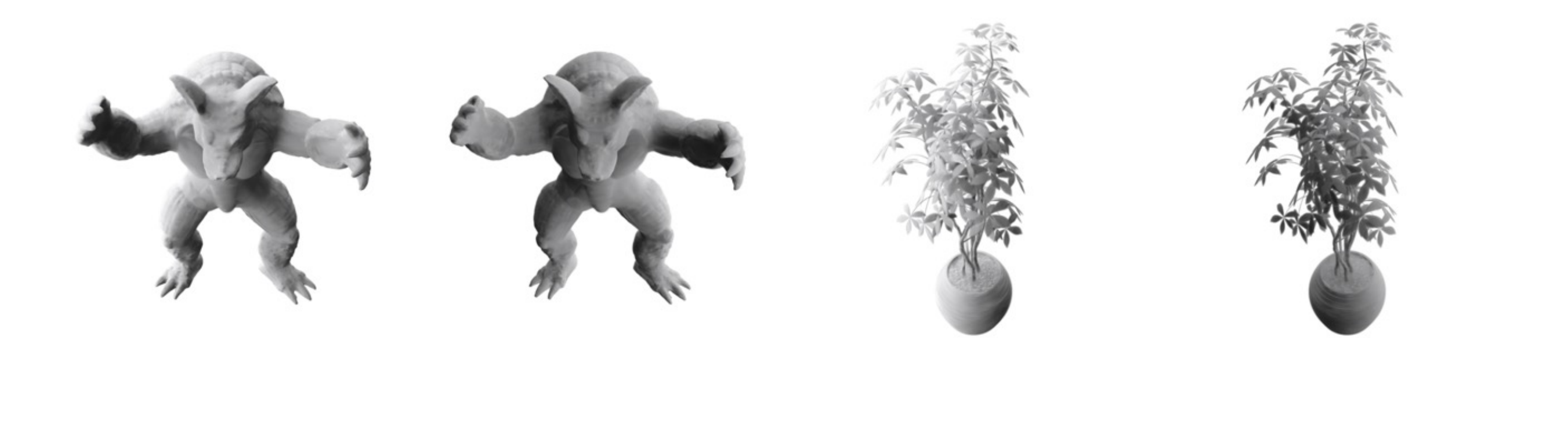}
\end{center}
\vspace{-20pt}
\caption{Visualization of diffuse light from a different point light.}
\vspace{-10pt}
\label{fig:point_light}
\end{figure}

\begin{figure}[t]
\begin{center}
\includegraphics[width=0.9\linewidth]{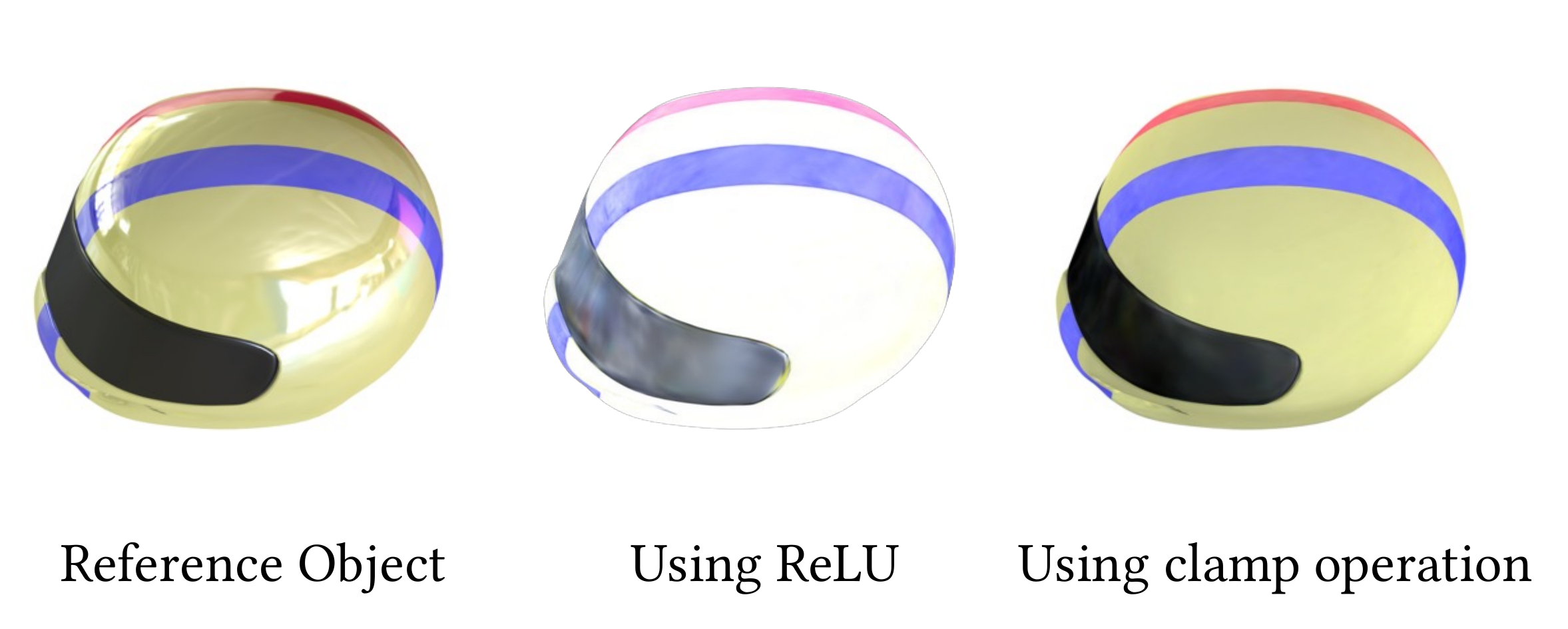}
\end{center}
\vspace{-18pt}
\caption{Albedo visualization for the effectiveness of albedo clamp operation.}
\vspace{-10pt}
\label{fig:clamp}
\end{figure}

\vspace{4pt}
\noindent \textbf{Implementation Details.} We also perform qualitative ablation studies to demonstrate the effectiveness of specific implementation details. We experiment with a ball object to clearly show how normal optimization is influenced by directional masking. As shown in Fig.~\ref{fig:dir_mask}, we first train two GIR models without directional masking for $4,999$ iterations. Subsequently, we activate the directional masking for one model and train both models for 15,000 iterations. It is evident that the GIR model trained with the proposed direction masking scheme exhibits higher accuracy in estimating normals than its counterpart. As discussed in the PBR estimation section (Sec.\ref{sec:method:normal_estimation}), we assigned random colors to the 3D Gaussians with back-view normals to distinguish masked Gaussians. Fig.~\ref{fig:maks_random_color} compares the performance with and without these strategies. Moreover, we sample 128 rays within a hemisphere for diffuse lighting to calculate occlusion for shadow computation, with the results illustrated in Fig.~\ref{fig:point_light}. To enhance the albedo estimation, we have replaced the ReLU activation with a clamp operation, constraining $\mathbf{a}$ within the range of $[0,1]$. The resulting improvements are demonstrated in Figure~\ref{fig:clamp}. Finally, to enhance the robustness of our method, we incorporated a random background during training, which contributes to better training of our GIR and improves overall performance.

\section{Discussion}
\label{sec:discussion}
Each of the works we compared demonstrates innovative and thoughtful designs, highlighting this research field's significance and evolving nature. In this section, we emphasize the distinctions between our proposed GIR method and the concurrent works, considering both technical design and demonstrated performance. By examining these aspects, we provide a comprehensive understanding of our approach's unique contributions and advantages.

% HDR
\begin{figure*}[t]
\begin{center}
\includegraphics[width=0.95\linewidth]{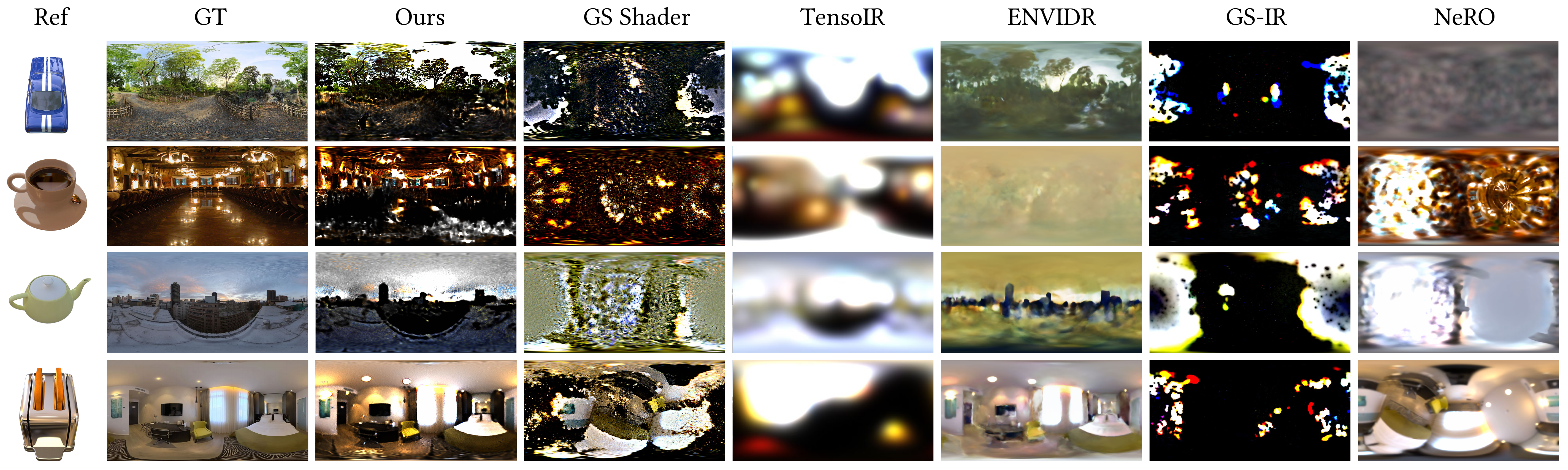}
\end{center}
    \vspace{-12pt}
   \caption{Qualitative comparison of HDR prediction on the Shiny Blender dataset. Our method exhibits high-quality HDR prediction, demonstrating the effectiveness of scene factorization.}
   \vspace{-10pt}
\label{fig:HDR}
\end{figure*}

\vspace{4pt}
\noindent \textbf{Relightable 3D Gaussian~\cite{gao2023relightable}} incorporates the neural light representation from NeIFL~\cite{yao2022neilf} into 3DGS. It introduces PBR properties and normal within each 3D Gaussian. Moreover, it proposes a geometry enhancement technique that utilizes depths predicted by a multi-view stereo (MVS) network and pseudo normals derived from depth maps under the assumption of local planarity. However, MVS net encounters challenges when dealing with glossy or reflective scenes, primarily due to the inherent difficulty of establishing accurate correspondences in such scenarios. In contrast, our proposed method utilizes the shortest eigenvector obtained from the covariance matrix of each 3D Gaussian to represent its normal. This approach acts as a self-supervised method without relying on external supervision with the help of our proposed directional masking module. Moreover, in the ray tracing module, their method suggests using a bounding volume hierarchy represented by a point cloud for efficient ray tracing. Our method proposes a grid-based tracing method specifically tailored for 3DGS, which offers efficiency advantages.

\vspace{4pt}
\noindent \textbf{GS-IR~\cite{liang2023gs}} is inspired by precomputation techniques used in the video game industry and adopts a baking-based occlusion approach to model indirect lighting. They utilize spherical harmonics to store occlusion information and represent indirect illumination. To obtain plausible normals, they introduce a strategy to enhance depth and utilize depth gradients to derive pseudo normals. However, this approach heavily relies on the quality of learned depth maps by 3DGS, unlike the Relightable 3D Gaussian~\cite{gao2023relightable} which uses MVS networks to predict depth maps as supervision. Obtaining accurate depth maps is challenging in 3DGS, particularly for complex scenes. This difficulty sometimes hampers the successful estimation of normals, leading to failures in the inverse rendering process.

\vspace{4pt}
\noindent \textbf{GaussianShader~\cite{jiang2023gaussianshader}} utilizes a low-resolution cubemap to represent lighting and employs the pre-integrated method for generating multiple mip maps. In contrast, our method introduces an FCN compression technique that allows for learning high-resolution environment maps. Additionally, they simplify the indirect illumination by incorporating a residual color term $\vc_r(\boldsymbol{\omega_o})$ stored within each 3D Gaussian, rather than tracing the outgoing radiance from other 3D Gaussians. In contrast, our method proposes an efficient grid-based tracing method to simulate the hemisphere integral of indirect illumination using a set of trainable spherical harmonics. Meanwhile, GaussianShader~\cite{jiang2023gaussianshader} uses the shortest axis as a prior and determines the normal direction by predicting residuals, without explicitly specifying the Gaussian orientations. It can result in each Gaussian being calculated in two directions, which poses challenges for thin objects and may lead to dual normals.

\vspace{4pt}
\noindent \textbf{Results.} GIR outperforms TensoIR~\cite{jin2023tensoir} in the relighting task, as measured by PSNR and LPIPS. The SSIM comparison between our method and TensoIR shows comparable results (0.943 vs. 0.945). These differences may stem from local relighting discrepancies in brightness, as Gaussian models tend to produce less continuous and smooth results compared to NeRF models. However, on the real-world OWL dataset, our method outperforms TensoIR. The varied performance across datasets can be attributed to differences in method design and rendering pipelines. For example, ENVIDR~\cite{liang2023envidr} estimates only specular and diffuse components without explicitly considering materials, performing well on datasets with metallic characteristics but less effectively on non-metallic ones. In contrast, TensoIR is better suited for non-metallic datasets with more diffuse components. Our approach, while not specifically tailored to any particular dataset, leverages a robust Gaussian representation and a more universal rendering pipeline, providing a competitive advantage in overall performance.

\vspace{4pt}
\noindent \textbf{Limitations.} 
While our method exhibits better performance compared to existing methods and holds potential for widespread application thanks to its real-time rendering capability using 3DGS representation, it still faces the challenge of albedo-illumination ambiguity, similar to other techniques. For instance, distinguishing between shadows cast by a fixed point light source caused by lighting or the albedo itself can be difficult. Additionally, our method utilizes pre-integrated lighting for PBR rendering, incorporating estimated indirect illumination through SH coefficients. Exploring a path-tracing approach specifically tailored for 3DGS representation would be an intriguing research direction to pursue in the future.

\section{Conclusion}
\label{sec:conclusion}
In conclusion, this paper introduces GIR, a novel method that utilizes 3DGS for scene factorization. GIR employs explicit 3D Gaussians to represent geometry and material properties, implementing BRDF rendering through splatting. The integration of inverse rendering into 3DGS faces three main challenges, which we address by directional masking scheme, effective illumination representation, and voxel-based tracing for indirect illumination. Through comprehensive experimentation, our approach demonstrates high performance compared to relevant approaches, with experimental results showcasing its capabilities in relighting and material editing.

\section*{Acknowledgments}
This work was supported in part by National Key Research and Development Program of China (2022YFF0904303), National Natural Science Foundation of China (61932003), and Beijing Science and Technology Planning Project (Z221100006322003). C. Wu is the project lead.

\bibliographystyle{IEEEtran}
\bibliography{sample-bibliography}

\end{document}